\documentclass[preprint,12pt]{elsarticle}

\usepackage{amssymb, amsfonts, amsmath}
\usepackage{soul}
\usepackage{xcolor}
\sethlcolor{gray!20} % Adjust the percentage for lighter or darker shades
\usepackage{hyperref}
\usepackage{listings}
\DeclareFixedFont{\ttb}{T1}{txtt}{bx}{n}{8} % for bold
\DeclareFixedFont{\ttm}{T1}{txtt}{m}{n}{8}  % for normal

\usepackage{color}
\usepackage{xcolor} % For defining colors

\definecolor{deepblue}{rgb}{0.1,0.1,0.8} % Keywords (e.g., functions, classes)
\definecolor{deepred}{rgb}{0.8,0.1,0.1}  % Custom/emphasized elements
\definecolor{deepgreen}{rgb}{0.1,0.6,0.1} % Strings
\definecolor{commentgray}{rgb}{0.5,0.5,0.5} % Comments
\definecolor{backgroundgray}{rgb}{1,1,1} % Background
\usepackage{listings}

\newcommand\pythonstyle{\lstset{
language=Python,
backgroundcolor=\color{backgroundgray}, % Set background color
basicstyle=\fontfamily{pcr}\selectfont\footnotesize, % Use Courier font with pdflatex
emph={MyClass,__init__,run_taqr, calculate_scores, simulate_wind_power_sde, run_nabqr_pipeline},       % Custom highlighted items (add functions here)
emphstyle=\bfseries\color{deepblue},     % Style for emphasized text
stringstyle=\color{deepgreen},          % Strings
commentstyle=\itshape\color{commentgray}, % Comments (italic + gray)
frame=tb,                           % Frame around code block
showstringspaces=false,                 % No visible spaces in strings
literate={->}{{$\to$}}1                 % Replace "->" with "→"
}}

\lstnewenvironment{python}[1][]
{
\pythonstyle
\lstset{#1}
}
{}

\newcommand\pythoninline[1]{{\pythonstyle\lstinline!#1!}}
\journal{SoftwareX}

\begin{document}
\renewcommand{\labelenumii}{\arabic{enumi}.\arabic{enumii}}

\begin{frontmatter}
 


\title{nabqr: Python package for improving probabilistic forecasts}
% nabqr: Python package for improving probabilistic forecasts
% nabqr: Python package for Neural Adaptive Basis for (time-adaptive) Quantile Regression

\affiliation[dtu]{organization={DTU Compute},%Department and Organization
            addressline={Bygning 303B}, 
            city={Kgs. Lyngby},
            postcode={2800}, 
            %state={},
            country={Denmark}}
            
 \affiliation[peter]{organization={QUENT ApS},
             addressline={Svanemøllevej 41},
             city={Hellerup},
             postcode={2900},
             country={Denmark}}
\author[dtu]{Bastian Schmidt Jørgensen\corref{cor1}}
% \ead{bassc@dtu.dk}
%\fntext[label2]{}
\cortext[cor1]{Corresponding author. \textit{Email address:} {bassc@dtu.dk}.}
\author[dtu]{Jan Kloppenborg Møller} 
\author[dtu,peter]{Peter Nystrup}
\author[dtu]{Henrik Madsen} 

\begin{abstract}
We introduce the open-source Python package NABQR: Neural Adaptive Basis for (time-adaptive) Quantile Regression that provides reliable probabilistic forecasts.
NABQR corrects ensembles (scenarios) with LSTM networks and then applies time-adaptive quantile regression to the corrected ensembles to obtain improved and more reliable forecasts. 
With the suggested package, accuracy improvements of up to 40\% in mean absolute terms can be achieved in day-ahead forecasting of onshore and offshore wind power production in Denmark. 
\end{abstract}

\begin{keyword}
probabilistic forecasting \sep LSTM networks \sep time adaptivity \sep quantile regression \sep  wind power production  



\end{keyword}

\end{frontmatter}

%\linenumbers

\section*{Metadata}
\label{}
Table \ref{codeMetadata}.
\begin{table*}[!h]
\centering
\begin{tabular}{|l|p{6.5cm}|p{6.5cm}|}
\hline
\textbf{Nr.} & \textbf{Code metadata description} & \textbf{Metadata} \\
\hline
C1 & Current code version & 0.1 \\
\hline
C2 & Permanent link to code/repository used for this code version & \url{https://github.com/bast0320/nabqr} \\
\hline
C3  & Permanent link to Reproducible Capsule & \url{https://github.com/bast0320/nabqr/blob/main/src/NABQR/nabqr.py}\\
\hline
C4 & Legal Code License   & MIT License \\
\hline
C5 & Code versioning system used & git, pypi \\
\hline
C6 & Software code languages, tools, and services used & Python, R \\
\hline
C7 & Compilation requirements, operating environments \& dependencies & python $\geq$ 3.10 and the following packages: 
icecream, matplotlib, numpy, pandas, properscoring, rich, SciencePlots, scikit\_learn, scipy, tensorflow, tensorflow\_probability, torch, typer, myst\_parser, tf\_keras

R with the following packages installed: quantreg, readr \\
\hline
C8 & If available Link to developer documentation/manual & For example: \url{https://nabqr.readthedocs.io/en/latest/} \\
\hline
C9 & Support email for questions & bassc\^*AT\^*dtu.dk \\
\hline
\end{tabular}
\caption{Code metadata}
\label{codeMetadata} 
\end{table*}

\section*{Abbreviations}
Table \ref{tab:abbreviations}.
\begin{table*}[h!]
    \centering
    \begin{tabular}{ll}
        \hline
        \textbf{Abbreviation} & \textbf{Description} \\
        \hline
        CRPS & Continuous Ranked Probability Score \\
        MAE & Mean Absolute Error \\
        QSS & Quantile Skill Score \\
        TAQR & Time-Adaptive Quantile Regression \\
        NABQR & Neural Adaptive Basis Quantile Regression \\
        CSV & Comma-Separated Values \\
        \hline
    \end{tabular}
    \caption{List of Abbreviations}
    \label{tab:abbreviations}
\end{table*}

\section{Motivation and significance}
Quantifying predictive uncertainty is a key challenge in many scientific fields that depend on model-based forecasts \citep{uncertQuant}. This is particularly important in areas such as energy systems, climate modeling, and environmental planning, where decisions must often be made under uncertainty and carry significant economic or operational consequences. Our software addresses this need by providing a streamlined and effective framework for converting ensemble forecasts—such as those generated by weather simulations from ECMWF \citep{ecmwf_website}—into accurate probabilistic forecasts.
%\red{uncertainty is important. examples} \\

Minimizing uncertainty is a universal goal in all forecasting domains, and our software package Neural Adaptive Basis for (time-adaptive) Quantile Regression, NABQR, provides a novel way to achieve it. Assume some (pre-existing) ensembles exist. That is, scenarios describing possible future developments. We wish to convert them into accurate quantile-based predictions. This process is broadly relevant, as such predictions can be applied to many areas, including risk management in finance, electricity markets, and production. We have applied NABQR to wind forecasting with strong results, demonstrating how refined ensemble forecasts translate into more reliable quantile estimates and improve forecast accuracy \citep{nabqr}.
%\red{everyone wants to improve forecasts...inherently} \\

Neural networks systematically correct ensemble forecasts, followed by time-adaptive quantile regression (TAQR) to estimate conditional quantile forecasts. This approach reduces quantile crossings and enhances forecast reliability. It supports scientific discovery by allowing users to incorporate data-driven corrections into their pipelines. Recent research on improving wind power prediction densities \citep{nabqr} shows how TAQR applied to refined ensembles leads to superior forecast distributions.
%\red{how it works briefly} \\

Using the software is straightforward: users supply ensemble forecasts and specify their target quantiles. The neural network component—drawing inspiration from architectures such as N-BEATS \citep{nbeats}—discovers basis functions that capture evolving data patterns, and an efficient TAQR solver \citep{taqrpaper} then produces calibrated quantiles. The open-source package, hosted on GitHub\footnote{\url{https://github.com/bast0320/nabqr}}, ensures transparency, reproducibility, and opportunities for community-driven enhancements. Its ease of integration facilitates adoption into broader workflows, enabling model comparison, scenario testing, and iterative improvements.
%\red{how to implement} \\

Our approach builds on and extends a rich body of literature. Quantile regression has a long history in statistical modeling \citep{qr_SOTA}, while neural networks have enabled flexible functional approximations \citep{nbeats,Cannon2011,Bremnes2020}. Ensemble-based methods from state-of-the-art weather forecasting centers \citep{ecmwf_website} naturally inform our inputs for probabilistic modeling. By combining ensemble calibration, neural network corrections, and time-adaptive quantile regression, our software creates a cohesive, methodologically sound, and practically useful framework. Its combination of neural network-based calibration, TAQR modeling, and user-friendly interfaces serves as a catalyst for advancing the state of the art in probabilistic forecasting and fostering scientific insights across diverse fields.
% \red{wrapping up...} \\

\section{Software description}
NABQR is an open-source Python package designed to enhance the accuracy and reliability of probabilistic forecasts. Built for flexibility and ease of integration, NABQR corrects ensemble forecasts using Long Short-Term Memory (LSTM) networks and applies Time-Adaptive Quantile Regression (TAQR) to produce optimal predictive distributions.

The software is implemented entirely in Python (compatible with version 3.10 and above). It adheres to modern software development practices, including modular design, comprehensive documentation, and integration with widely-used libraries such as TensorFlow, NumPy, and pandas. NABQR supports various input formats and produces results in accessible formats like CSV and NumPy arrays, facilitating seamless incorporation into existing workflows.

NABQR provides users with a streamlined pipeline for processing ensemble data, training neural networks, and applying quantile regression methods.

For comprehensive and well-documented examples, users are encouraged to visit the following ReadTheDocs\footnote{\url{https://nabqr.readthedocs.io/en/latest}}. The documentation describes its usage in full detail.

NABQR has been tailored specifically to work well with wind power production forecasting in Denmark. Running the software with almost no tuning provides upwards of 40\% improvements in the mean absolute sense for specific areas.

\subsection{Software architecture}
NABQR's Application Programming Interface (API) is designed to be compatible with the widely-used TensorFlow library \citep{tensorflow}, facilitating seamless integration into existing machine learning workflows. The package supports established data formats, such as NumPy arrays \citep{numpy} and pandas data frames \citep{pandas}, ensuring ease of use and interoperability. Data generated by NABQR can be saved in pickle, NumPy, or CSV formats, allowing for flexible data handling and inspection.

To enhance extensibility and maintainability, NABQR employs object-oriented design principles, particularly when dealing with neural networks. This approach modularizes the framework, with each component encapsulated within a class or function. The repositories adhere to the Black\footnote{\url{https://github.com/psf/black}} coding style to ensure consistency and enhance readability.

Additionally, NABQR includes a comprehensive Python module that implements TAQR, featuring user-friendly function wrappers. The framework's components are integrated into a central \hl{Pipeline} function wrapper, providing users with access to a variety of convenient methods, including logging and visualization capabilities, see Fig. \ref{fig:pipeline_flowchart}. NABQR can be easily incorporated into Python scripts or Jupyter Notebooks, supporting tasks such as training LSTM networks, executing TAQR algorithms, or running the entire pipeline.
NABQR primarily consists of 5 Python modules:
\begin{itemize}
    \item \textbf{functions\_for\_TAQR.py}:
    This module contains the main functions of TAQR. It includes algorithms for updating and solving quantile regression problems using the simplex method.

    \item \textbf{functions.py}:
    This module provides core functionalities for the NABQR framework, including scoring metrics such as Variogram, CRPS, MAE and QSS. It also includes functions for dataset creation, preprocessing, and model definitions, as well as the implementation of the TAQR algorithm.

    \item \textbf{helper\_functions.py}:
    This module offers utility functions that support the main operations of the NABQR framework. It includes functions for data manipulation, and simulating correlated AR(1) processes as well as more advanced and realistic wind power production (see \texttt{simulate\_wind\_power\_sde}), which is used for generating synthetic datasets and testing robustness.

    \item \textbf{nabqr.py}:
    This is the main interface for running the NABQR pipeline. It integrates ensemble generation, model training, and visualization capabilities, providing a comprehensive pipeline for executing the NABQR methodology from data simulation to result visualization. This is also where simulated data can be generated.

    \item \textbf{visualization.py}:
    This module is responsible for visualizing the results of the NABQR models. It includes functions to generate plots that illustrate the performance of the quantile regression models, helping users interpret and analyze the probabilistic forecasts effectively and ensure consistency in plotting, for example using functionalities from \hl{matplotlib.mdates.AutoDateLocator} and a dynamic way to plot the number of ensembles.
\end{itemize}

The code is well-documented and easy to navigate, and people are welcome to make changes to their copy of the package under the MIT license.

\subsection{Software functionalities}
NABQR trains an LSTM network to correct the ensembles, to be more reliable, and to have a better quantile score. We optimize for a multidimensional quantile loss function; that is, we optimize for all quantiles in the same network, and not a separate network for each quantile. The output of the LSTM network is used as a basis matrix for a time-adaptive quantile regression algorithm, which produces probabilistic multivariate forecasts. Probabilistic forecast measures have been implemented to readily calculate a range of scores: MAE, CRPS, Variogram-Score, Quantile Score, and reliability plots. 

To install the package run: \\
{\footnotesize $\gg$   \normalsize \texttt{  pip install nabqr}}

\subsubsection{Submodules}
NABQR provides practical and useful wrapper functions. We will use the notation \verb|function()|$\to$\verb|(output1, ...)| to explain the following functions. For more information, see the documentation. \\

\begin{python}
calculate_scores() -> (MAE,QS,CRPS,VarS)
\end{python}
When provided with the observations, TAQR predictions, and ensembles, this function will output MAE, QS, CRPS, and VarS scores to evaluate the probabilistic forecast's performance.

\begin{python}
simulate_wind_power_sde() -> (ensembles,y)
\end{python}
Simulates normalized wind power production using an Ornstein-Uhlenbeck process with GARCH volatility and jumps of normally distributed sizes driven by a Poisson process. The mean reversion is state-dependent with a repelling mechanism near 1.0 (upper boundary), and the diffusion term vanishes at the boundaries to avoid unphysical values outside [0, 1]. Returns ensembles and observations ($y$).

\begin{python}
run_taqr(X, y, q_list, n_init, n_full) -> (q_hat, y, beta)
\end{python}
Executes the full TAQR algorithm from timestep $t_{n_{\text{init}}}$ to $t_{n_{\text{full}}}$ \citep{taqrpaper}. It employs \small \verb|one_step_quantile_prediction()|, \normalsize which can run the TAQR algorithm for at least one timestep and up to all available timesteps for a single quantile. Returns the predicted quantiles $\hat{q}$, observations, $y$, for convenience, and the time-adaptive parameters, $\beta$.

\begin{figure*}
    \centering
    \includegraphics[width=0.7\linewidth]{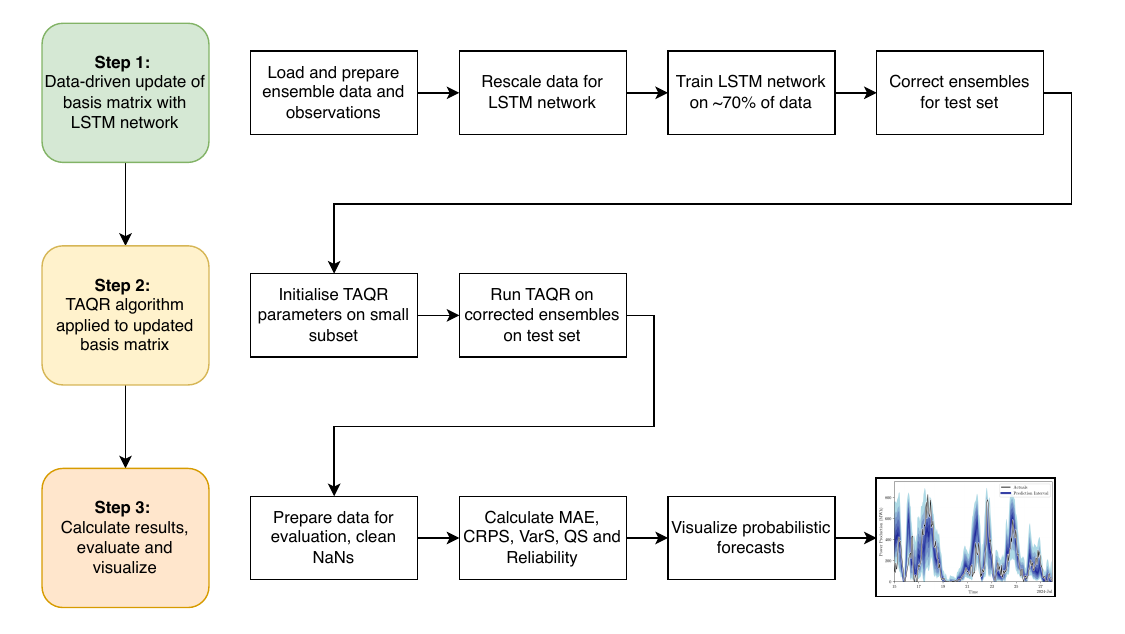}
    \caption{Overview of pipeline in NABQR. Note that for the first step if data is not specified to be loaded the pipeline will automatically simulate data. }
    \label{fig:pipeline_flowchart}
\end{figure*}

\begin{python}
run_nabqr_pipeline(X,y, ...) -> (scores, corr_ens, q_hat, ...) 
\end{python}
Fig. \ref{fig:pipeline_flowchart} shows the step in the pipeline function. The workflow involves loading and preparing ensemble data and observations, followed by training an LSTM network on approximately 70\% of the data to update the basis matrix. Once the data is corrected, the TAQR algorithm is initialized on a small subset and applied to the test set. The results are then prepared by cleaning missing values (NaNs) and evaluating MAE, CRPS, VarS, QS, and reliability. Finally, probabilistic forecasts are visualized for interpretation and analysis. Returns the calculated scores, corrected ensembles, $\hat{q}$. For the full output and the files it saves please see the documentation.

% Inspiration: https://www.sciencedirect.com/science/article/pii/S2352711024003443

\section{Illustrative examples}
\begin{figure}[!ht]
    \centering
    \includegraphics[width=1\linewidth]{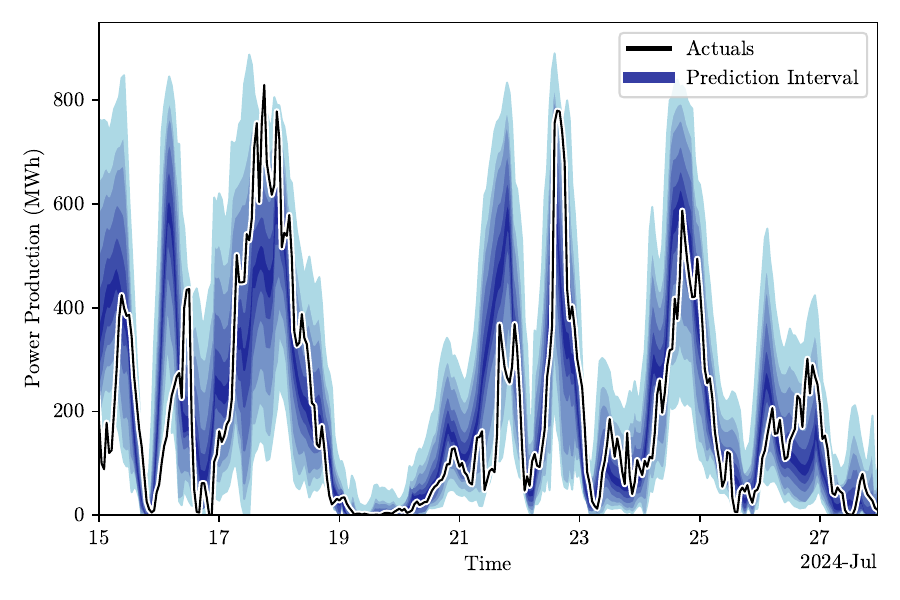}
    \caption{Excerpt of TAQR predictions with prediction interval 5\%-95\% shown in shades of blue for Denmark (area DK2) Offshore wind power production. Obtained by running \texttt{run\_nabqr\_pipeline} on wind power production.}
    \label{fig:real_output}
\end{figure}

Running the \verb|run_nabqr_pipeline| with the only inputs being wind power production data, \verb|(X=ensembles,y=observations)|, will create Fig. \ref{fig:real_output} as in \citep{nabqr}.

Initially, the code imports essential components, including NABQR helper functions, the \hl{SciencePlots} plotting library, and the \hl{datetime} library. Subsequently, a synthetic dataset is generated for testing purposes. An example of using the \texttt{simulate\_wind\_power\_sde}, can be seen in \ref{app:simulation}.
The final section of the code executes the entire pipeline, specifying the dataset, the training size for the LSTM network, the time steps (representing the lagged values for the LSTM network), and the quantiles for the TAQR algorithm. 
Fig. \ref{fig:real_output} shows predictions for actual wind power production, see \citep{nabqr}.

\section{Impact}
The NABQR package improves renewable energy power production forecasts, achieving up to 40\% better accuracy in terms of mean absolute error compared to previous state-of-the-art commercial forecasts. This directly impacts the efficiency and sustainability of power grid operations. By improving forecast accuracy, NABQR reduces reliance on backup power plants, thereby decreasing CO2 emissions and supporting the green transition. This advancement encourages investment and innovation in renewable energy and accelerates the transition to sustainable energy sources.

NABQR democratizes advanced machine learning techniques, making them accessible to a wider audience beyond domain experts. This aligns with the principles of human-centered AI, promoting transparency, collaboration, and inclusivity. By simplifying machine learning pipelines, NABQR reduces the need for manual trial-and-error, streamlining the knowledge discovery process and empowering users to efficiently manage and analyze datasets.

%The package's integration into a leading energy forecasting provider's pipeline by 2025 underscores its practical utility and potential for widespread adoption. 
Written in Python and thoroughly documented, NABQR is set to be used in both academic and industrial settings. This accessibility invites further exploration of neural networks as optimal basis function generators, fostering new research opportunities and advancements in the field.

\section{Conclusions}
NABQR is an open-source Python package designed to correct scenarios and produce more accurate reliable probabilistic forecasts. NABQR uses neural networks and time-adaptive quantile regression. The NABQR package includes the first time-adaptive quantile regression algorithm written in Python along with useful wrapper functions for e.g. visualizing probabilistic forecasts, training LSTM networks, and simulating realistic wind power.

\subsection*{CRediT authorship contribution statement}
\textbf{Bastian Schmidt Jørgensen:} Writing– original draft, Project administration, Investigation, Data curation, Software, Methodology. 

\textbf{Jan Kloppenborg Møller:} Writing– review \& editing, Conceptualization, Software. 

\textbf{Peter Nystrup}:
Writing– review \& editing, Data curation, Conceptualization. 

\textbf{Henrik Madsen:} Writing– review \& editing, Conceptualization, Funding, Project administration.

\subsection*{Declaration of competing interest}
The authors declare that they have no known competing financial
interests or personal relationships that could have appeared to influence the work reported in this article.
\subsection*{Data availability}
The authors do not have permission to share data, but simulated data is available in NABQR.
\subsection*{Acknowledgements}
This work is supported by \textit{IIRES} (Energy Cluster Denmark),  \textit{SEM4Cities} (Innovation Fund Denmark, No. 0143-0004), \textit{ELEXIA} (Horizon Europe No. 101075656), and the projects \textit{DynFlex} and \textit{PtXMarkets}, which both are part of the Danish Mission Green Fuel portfolio of projects. 

\section*{Personal Acknowledgements}
\label{}
Thanks to Peder Bacher and Goran Goranovic for good discussion.

\newpage
\bibliographystyle{elsarticle-num-names} 
\bibliography{refs.bib}
\appendix
\section{Example of Simulated Data}
\label{app:simulation}
\begin{figure}[!htb]
    \centering
    \includegraphics[width=1\linewidth]{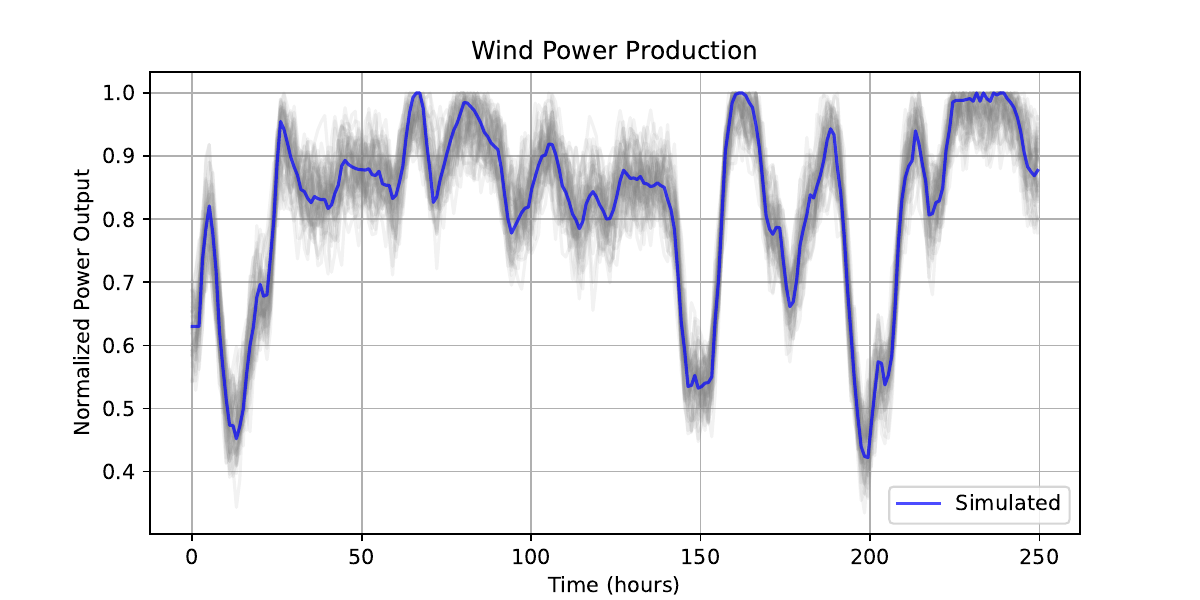}
    \caption{Simulation for 250 hours of normalized wind power data including ensembles with the method \texttt{simulate\_wind\_power\_sde} in \texttt{run\_nabqr\_pipeline}.}
    \label{fig:app_sim}
\end{figure}

\end{document}